\def\year{2019}\relax
\documentclass[letterpaper]{article} 
\usepackage{aaai19}  
\usepackage{times}  
\usepackage{helvet}  
\usepackage{courier}  
\usepackage{url}  
\usepackage{graphicx}  
\usepackage{amsfonts}
\usepackage{amsmath}
\usepackage{multirow}
\usepackage{booktabs}
\usepackage{url}
\usepackage{amsthm}
\usepackage{tabularx}
\usepackage{xcolor}
\usepackage{tcolorbox}
\usepackage[linesnumbered,ruled]{algorithm2e}
\usepackage{footnote}
\makesavenoteenv{tabular}
\makesavenoteenv{table}
\theoremstyle{definition}
\newtheorem{defn}{Definition}[]

\newtcbox{\xmybox}[1][red]{on line,
arc=0pt,outer arc=0pt,colback=#1!10!white,colframe=#1!100!black,
boxsep=0pt,left=1pt,right=1pt,top=2pt,bottom=2pt,
boxrule=0pt,bottomrule=1pt,toprule=1pt}

\begin{document}
%
\title{A Hierarchical Framework for Relation Extraction \\
    with Reinforcement Learning}
\author{Ryuichi Takanobu$^{1,3}$\footnotemark[1],
        Tianyang Zhang$^{1,3}$\footnotemark[1],
        Jiexi Liu$^{2,3}$\footnotemark[1],
        Minlie Huang$^{1,3}$\footnotemark[2]\\
  $^1$ Dept. of Computer Science \& Technology, $^2$ Dept. of Physics, Tsinghua University, Beijing, China \\ 
  $^3$ Institute for Artificial Intelligence, Tsinghua University (THUAI), China \\
  $^3$ Beijing National Research Center for Information Science \& Technology, China \\
  {\tt \{gxly15, zhang-ty15, liujx15\}@mails.tsinghua.edu.cn, aihuang@tsinghua.edu.cn}\\
}
\maketitle
\renewcommand{\thefootnote}{\fnsymbol{footnote}}
\footnotetext[1]{All authors contributed equally to this work.}
\footnotetext[2]{Corresponding author: Minlie Huang.}
\renewcommand{\thefootnote}{\arabic{footnote}}
\begin{abstract}
  Most existing methods determine relation types only after all the entities have been recognized, thus the interaction between relation types and entity mentions is not fully modeled. This paper presents a novel paradigm to deal with relation extraction by regarding the related entities as the arguments of a relation. We apply a hierarchical reinforcement learning (HRL) framework in this paradigm to enhance the interaction between entity mentions and relation types. The whole extraction process is decomposed into a hierarchy of two-level RL policies for relation detection and entity extraction respectively, so that it is more feasible and natural to deal with overlapping relations. Our model was evaluated on public datasets collected via distant supervision, and results show that it gains better performance than existing methods and is more powerful for extracting overlapping relations\footnote{Data and code are publicly available at: \url{https://github.com/truthless11/HRL-RE}.}.
\end{abstract}

\section{Introduction}

\begin{figure*}[!tp]
    \centering
    \includegraphics[width=\linewidth]{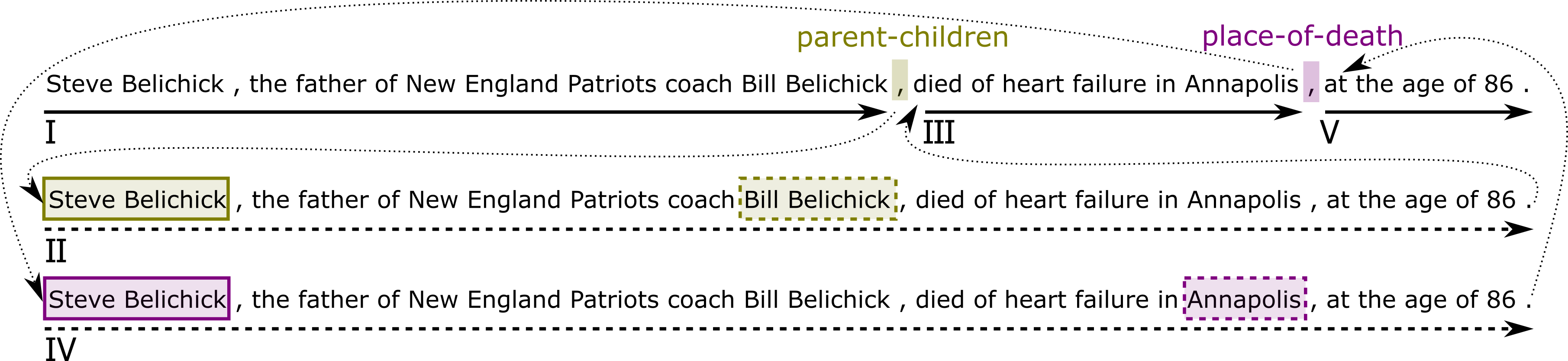}
    \caption{An example sentence which has two {\bf overlapping relations} (\textit{Steve Belichick}, \texttt{parent-children}, \textit{Bill Belichick}), (\textit{Steve Belichick}, \texttt{place-of-death}, \textit{Annapolis}). The solid arrow indicates the high-level relation detection process, and the dashed arrow for low-level entity extraction. The dotted arrow marks a transition between the two processes. This example shows how overlapping relations are extracted (\textit{Steve Blichick} is included in both triples).}
    \label{Example}
\end{figure*}

Extracting entities, relations, or events from unstructured texts is crucial for building large-scale, reusable knowledge which can facilitate many other tasks \cite{mintz2009distant,nadeau2007survey}, including knowledge base construction \cite{dong2014knowledge,luan2018multi}, question answering \cite{fader2014open}, and biomedical text mining \cite{huang2015community}. 


The task of relation extraction is to identify relations $(e_s, r, e_t)$\footnote{Throughout this paper, a relation refers to a triple $(e_s, r, e_t)$, a relation type refers to $r$.}, a triple consisting of a relation type $r$, a source entity $e_s$ and a target entity $e_t$. In this paper, we propose a novel joint extraction paradigm in the framework of hierarchical reinforcement learning \cite{sutton1999between}, where we first detect a relation and then extract the corresponding entities as the argument of a relation.

Our model detects {\bf relation indicators} by a high-level reinforcement learning (RL) process and identifies the participating entities for the relation by a low-level RL process.  
As shown in Figure \ref{Example},
the extraction process makes sequential scans from the beginning to the end of a sentence (\textbf{I}). The high-level process is to detect a relation indicator at some particular position. 
If a certain relation is identified, a low-level sequential process is triggered to identify the corresponding entities for that relation (\textbf{II}). 
When the low-level subtask for entity extraction is completed (\textbf{III}), the high-level RL process continues its scan to search for the next relation (\textbf{IV}) in the sentence.

This paradigm has strengths in dealing with two issues existing in prior studies. 
\textbf{First}, most traditional models \cite{gormley2015improved,hoffmann2011knowledge,miwa2016end} determine a relation type only after all the entities have been recognized, whereas the interaction between the two tasks is not fully captured. In some sense, these methods are aligning a relation to entity pairs, and therefore, they may introduce additional \textit{noise} since a sentence containing an entity pair may not truly mention the relation \cite{zhang2013towards}, or may describe multiple relations \cite{takamatsu2012reducing}. 

\textbf{Second}, there still lacks the elegance of the joint extraction method to deal with {\it one-to-many problems} (\textbf{overlapping relations}): one entity may participate in multiple relations in the same sentence  (see \textit{Steve Blichick} in Figure \ref{Example}), or even the same entity pair within a sentence is associated with different relations. 
To our best knowledge, CopyR \cite{zeng2018extracting} is the only method that discussed this issue, which views relation extraction as a triple generation process. However, this method, as our experiments reveal, strongly relies on the training data, and cannot extract multi-word entity mentions. 

In our paradigm, the \textbf{first issue} is handled by treating entities as the arguments of a relation.
The dependency between entity mentions and relation types is formulated through designing the state representations and rewards in the high-level and low-level RL processes.
The interaction is well captured since the main task (high-level RL process for relation detection) passes messages when launching a subtask (low-level RL process for entity extraction), and the low-level rewards, signifying how well a subtask is completed, are passed back to the main task. In this manner, the interaction between relation types and entity mentions can be better modeled.

The \textbf{second issue} is addressed by our hierarchical structure. By decomposing relation extraction into a high-level task for relation detection and a low-level task for entity extraction, multiple relations in a sentence can be handled separately and sequentially. As shown in Figure \ref{Example}, the first relation is extracted when the main task detects the first relation type (\texttt{parent-children}), and the second relation is subsequently extracted when the second relation type (\texttt{place-of-death}) is triggered, even though the two relations share the same entity (\textit{Steve Blichick}). Experiments demonstrate the proposed paradigm achieves strong performance over the baselines in extracting overlapping relations.

In summary, our contributions are in two folds:
\begin{itemize}

\item 
We design a novel end-to-end hierarchical paradigm to jointly identify entity mentions and relation types, which decomposes the task into a high-level task for relation detection and a low-level task for entity extraction. 

\item 
By incorporating reinforcement learning into this paradigm, the proposed method outperforms baselines in modeling the interactions between the two tasks, and extracting overlapping relations.
\end{itemize}



\section{Related Work}

Traditional pipelined approaches treat entity extraction and relation classification as two separate tasks \cite{mintz2009distant,gormley2015improved,tang2015line}. They first extract the token spans in the text to detect entity mentions, and then discover the relational structures between entity mentions. Although it is flexible to build pipelined methods, these methods suffer from \textit{error propagation} since downstream modules are largely affected by the errors introduced by upstream modules. 

To address this problem, a variety of joint learning methods was proposed. \citeauthor{kate2010joint} \shortcite{kate2010joint} proposed a card-pyramid graph structure for joint extraction, and \citeauthor{hoffmann2011knowledge} \shortcite{hoffmann2011knowledge} developed graph-based multi-instance learning algorithms. However, the two methods both applied a greedy search strategy to reduce the exploration space aggressively, which limits the performance. Other studies employed a structured learning approach \cite{li2014incremental,miwa2014modeling}. All these models depend on \textit{heavy feature engineering}, which requires much manual efforts and domain expertise. 

On the other hand, \citeauthor{bjorne2011extracting} \shortcite{bjorne2011extracting} proposed to first extract \textbf{relation triggers}, which refer to a phrase that \textbf{explicitly} expresses the occurrence of a relation in a sentence, and then determine their arguments to reduce the task complexity. Open IE systems \textit{ReVerb} \cite{fader2011identifying} identifies relational phrases using lexical constraints, which also follows a ``relation"-first, ``argument"-second approach. But there are many cases where no relation trigger appears in a sentence so that such relations cannot be captured in these methods. 

Neural models for joint relation extraction are investigated in recent studies \cite{katiyar2016investigating,zhang2017end}. \citeauthor{miwa2016end} \shortcite{miwa2016end} proposed a neural model that shares parameters for entity extraction and relation classification, but the two tasks are separately handled, and the final decision is obtained via exhaustively enumerating the combinations between detected entity mentions and relation types. Unlike aforementioned methods that all the entities are recognized first, \citeauthor{zheng2017joint} \shortcite{zheng2017joint} used a tagging scheme which applies a Cartesian product of the relation type tags and the entity mention tags, and thus each word is assigned a unique tag that encodes entity mentions and relation types simultaneously. However, it is unable to deal with overlapping relations in a sentence: if an entity is the argument of multiple relations, the tag for the entity should not be unique.  
The recent study \cite{zeng2018extracting} is closely related to ours that aims to handle overlapping relations. It employs multiple decoders based on sequence-to-sequence (Seq2Seq) learning where a decoder copies an entity word from the source sentence and each triple in a sentence is generated by different decoders, but such a method strongly relies on the annotation of training data and it cannot extract an entity that has multiple words.

Reinforcement learning has been witnessed in information extraction very recently. RL was employed to acquire and incorporate external evidence in event extraction \cite{narasimhan2016improving}. \citeauthor{feng2018reinforcement} \shortcite{feng2018reinforcement} used RL to train an instance selector to denoise training data obtained via distant supervision for relation classification. 
Improvement was reported in distant supervision relation type extraction by exploring RL to redistribute false positives into the negative examples \cite{qin2018robust}.

\section{Hierarchical Extraction Framework}
 

\subsection{Overview}

First of all, we define \textit{relation indicator} as follows:
\begin{defn}
    \textit{Relation indicator} is the position in a sentence when sufficient information has been mentioned to identify a semantic relation. Different from \textit{relation trigger} (i.e., explicit relation mention), relation indicators can be verbs (e.g. \textit{die} of), nouns (e.g. his \textit{father}), or even prepositions (e.g. \textit{from/by}), other symbols such as comma and period (As shown in Figure \ref{Example}, the relation type \texttt{place-of-death} can be signified till the {\it comma} position).
\label{define:relation_indicator}  
\end{defn}
Relation indicator is crucial for our model to complete the extraction task, because the entire extraction task is decomposed into relation indicator detection and entity mention extraction.

\begin{figure}[!htp]
    \centering
    \includegraphics[width=\linewidth]{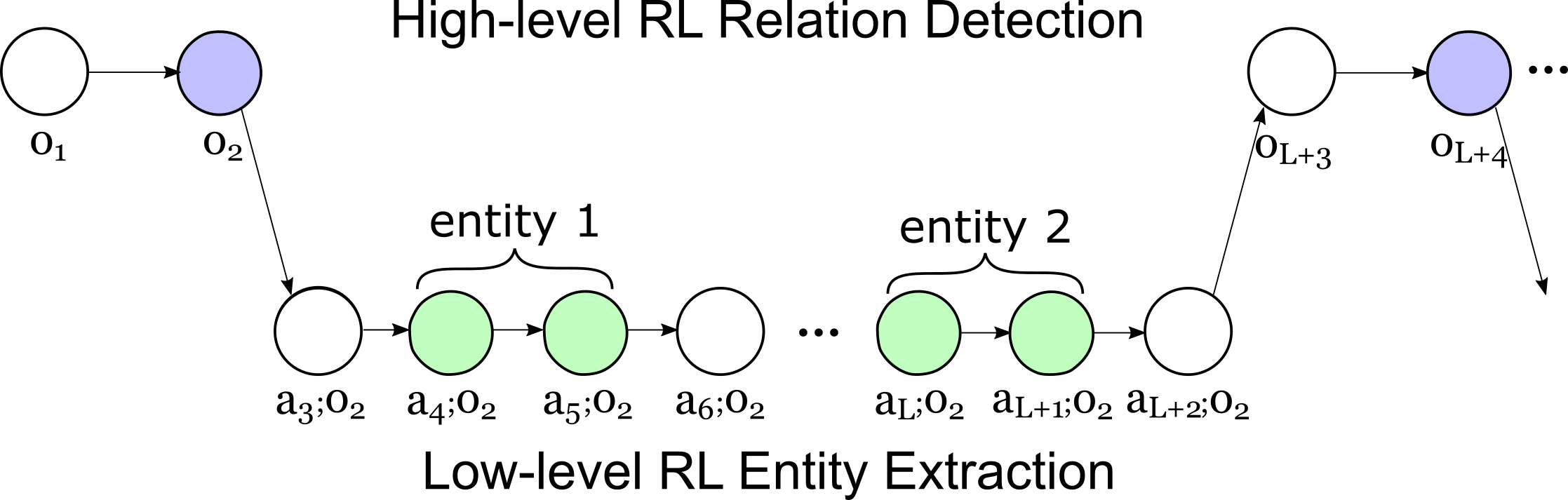}
    \caption{Overview of a hierarchical agent in relation extraction.} 
    \label{hierpolicy}
\end{figure}

The entire extraction process works as follows. An agent predicts a relation type at a particular position when it scans a sentence sequentially. Note that this process of relation detection needs no annotation of entities, thus different from relation classification which is to identify the relations between pairs of entities. When there is no sufficient evidence to indicate a semantic relation at a time step, the agent may choose \texttt{NR}, a special relation type that indicates no relation. Otherwise a relation indicator is triggered, the agent launches a subtask for entity extraction to identify the arguments of the relation, the two entities. When the entity mentions are identified, the subtask is completed and the agent continues to scan the rest of the sentence for other relations. 

Such a process can be naturally formulated as a semi-Markov decision process \cite{sutton1999between}: 1) a high-level RL process that detects a relation indicator in a sentence; 2) a low-level RL process that identifies the associated entities for the corresponding relation. 
By decomposing the task into a hierarchy of two RL processes, the model is advantageous at dealing with sentences which have multiple relation types for the same entity pair, or one-to-many entities in which an entity is the argument of multiple relations. 


\begin{figure}[!hptb]
    \centering
    \includegraphics[width=\linewidth]{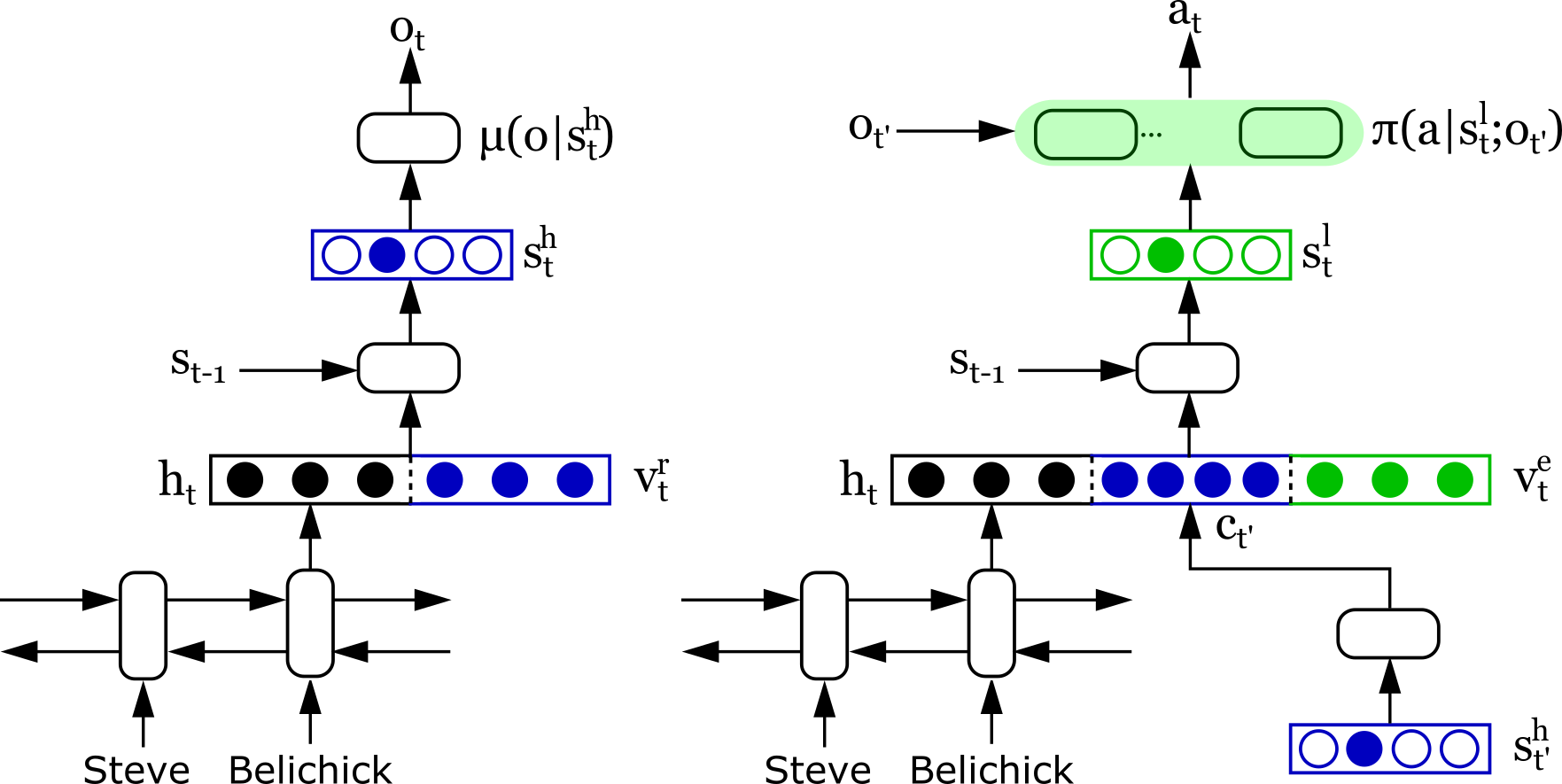}
    \caption{Illustration of a two-level hierarchical policy structure. Left panel shows the high-level policy for relation detection, and right panel shows the low-level policy for entity extraction.}
    \label{policy}
\end{figure}

\subsection{Relation Detection with High-level RL}

\begin{figure*}[!tpb]
    \centering
    \includegraphics[width=\linewidth]{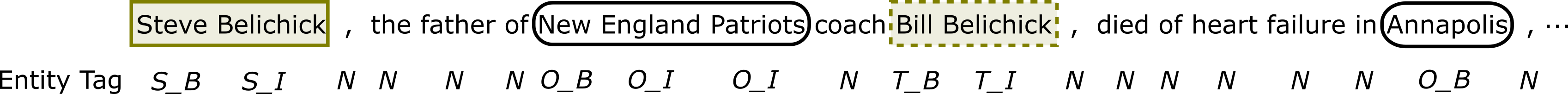}
    \caption{The entity annotation scheme for the example sentence in Figure \ref{Example} when the agent predicts a relation type \texttt{parent-children} between \textit{Steve Belichick} and \textit{Bill Belichick}. In this example, \textit{New England Patriots} and \textit{Annapolis} are not-concerned entities with respect to relation type \texttt{parent-children}.
    }
    \label{tag}
\end{figure*}

The high-level RL policy $\mu$ aims to detect the relations in a sentence $S = w_1 w_2 \cdots w_L$, which can be regarded as a conventional RL policy over options. An option refers to a high-level action, and a low-level RL process will be launched once an option is executed by the agent. 
\\
\textbf{Option}: The option $o_t$ is selected from $\mathcal{O} = \{ \texttt{NR} \} \cup \mathcal{R}$ where $\texttt{NR}$ indicates no relation, and $\mathcal{R}$ is the relation type set. When a low-level RL process enters a terminal state, the control of the agent will be taken over to the high-level RL process to execute the next options. 
\\
\textbf{State}: The state $\mathbf{s}_t^h$ $\in \mathcal{S}$ of the high level RL process at time step $t$, is represented by: 1) the current hidden state $\mathbf{h}_t$, 2) the relation type vector $\mathbf{v}_t^r$ (the embedding of the \textit{latest} option $o_{t'}$ that $o_{t'} \not= \texttt{NR}$, a learnable parameter), and 3) the state from the last time step $\mathbf{s}_{t-1}$\footnote{where $\mathbf{s}_{t-1} = \mathbf{s}_{t-1}^h$ if the agent sampled a high-level option at last time step $t-1$, and $\mathbf{s}_{t-1} = \mathbf{s}_{t-1}^l$ if the agent sampled a low-level action.}, formally represented by 
\begin{equation}\label{state_high}
    \begin{split}
        \mathbf{s}_t^h &= f^h(\mathbf{W}_s^h [ \mathbf{h}_t ; \mathbf{v}_t^r ; \mathbf{s}_{t-1}]),
    \end{split}
\end{equation}
where $f^h(\cdot)$ is a non-linear function implemented by MLP. To obtain the hidden state $\mathbf{h}_t$, we introduce a sequence Bi-LSTM over the current input word embedding $\mathbf{w}_t$:
\begin{equation}\label{LSTM}
    \begin{split}
        \overrightarrow{\mathbf{h}_t} &= \overrightarrow{LSTM}(\overrightarrow{\mathbf{h}_{t-1}}, \mathbf{w}_t), \\
        \overleftarrow{\mathbf{h}_t} &= \overleftarrow{LSTM}(\overleftarrow{\mathbf{h}_{t+1}}, \mathbf{w}_t), \\
        \mathbf{h}_t &= [ \overrightarrow{\mathbf{h}_t} ; \overleftarrow{\mathbf{h}_t} ].
    \end{split}
\end{equation}
\\
\textbf{Policy}: The stochastic policy for relation detection $\mu: \mathcal{S} \to \mathcal{O}$ which specifies a probability distribution over options:
\begin{equation}\label{option}
    o_t \sim \mu(o_t|\mathbf{s}_t^h) = softmax(\mathbf{W}_\mu \mathbf{s}_t^h).
\end{equation}
\textbf{Reward}: Then, the environment provides intermediate reward $r_t^h$ to estimate the future return when executing $o_t$. The reward is computed as below:
\begin{equation}\label{reward_high}
    r_t^h = \left\{
    \begin{array}{ll}
        -1, &if ~ o_t ~ not ~ in ~ S\\
        0, &if ~ o_t = \texttt{NR} \\
        1, &if ~ o_t ~ in ~ S.
    \end{array}
    \right.
\end{equation}
If $o_t = \texttt{NR}$ at certain time step, the agent transfers to a new high-level inter-option state at the next time step. Otherwise the low-level policy will execute the entity extraction process. The inter-option state will not transfer until the subtask over current option $o_t$ is done, which may take multiple time steps. Such a semi-Markov process continues until the last option about the last word $w_L$ of $S$ is sampled. Finally, a final reward $r_{fin}^h$ is obtained to measure the sentence-level extraction performance that $\mu$ detects:
\begin{equation}\label{reward_h_final}
    r_{fin}^h = F_\beta(S) = \frac{(1 + \beta^2) Prec \cdot Rec}{\beta^2 Prec + Rec},
\end{equation}
where F$_\beta$ is the weighted harmonic mean of {\it precision} and {\it recall} in terms of the relations in $S$. $Prec$/$Rec$ indicates precision/recall respectively, computed over one sentence.

\subsection{Entity Extraction with Low-level RL}
Once the high-level policy has predicted a non-$\texttt{NR}$ relation type, the low-level policy $\pi$ will extract the participating entities for the corresponding relation. The low-level policy over actions (primitive actions) is formulated very similarly as the high-level policy over options. To make the predicted relation type accessible in the low-level process, the option $o_{t'}$ from the high level RL is taken as additional input throughout the low-level extraction process. 
\\
\textbf{Action}: The action at each time step is to assign an entity tag to the current word. The action space, i.e., entity tag space $\mathcal{A} = (\{\texttt{S}, \texttt{T}, \texttt{O}\} \times \{\texttt{B}, \texttt{I}\}) \cup \{\texttt{N}\}$, where \texttt{S} represents the participating source entity, \texttt{T} for the target one, \texttt{O} for the entities that are not associated with the predicted relation type $o_{t'}$, and \texttt{N} for for non-entity words. Note that, the same entity mention may be assigned with different \texttt{S/T/O} tags depending on different relation types concerned at the moment. In this way, the model can deal with overlapping relations.
In addition, we use the \texttt{B/I} symbols to represent the beginning word and the inside of an entity, respectively. Refer to Figure \ref{tag} for an example. 
\\
\textbf{State}: Similar to the policy for relation detection, the low-level intra-option state $\mathbf{s}_t^l$ is represented by 1) the hidden state $\mathbf{h}_t$ of current word embedding $\mathbf{w}_t$, 2) the entity tag vector $\mathbf{v}_t^e$ which is a learnable embedding of $a_{t-1}$, 3) the state from previous time step $\mathbf{s}_{t-1}$, and 4) the context vector $\mathbf{c}_{t'}$ using the relational state representation assigned to the latest option $\mathbf{s}_{t'}^h$ in Eq. (\ref{state_high}), as follows:
\begin{equation}\label{state_low}
    \begin{split}
        \mathbf{c}_{t'} &= g(\mathbf{W}_h^l \mathbf{s}_{t'}^h) \\
        \mathbf{s}_t^l &= f^l(\mathbf{W}_s^l [ \mathbf{h}_t ; \mathbf{v}_t^e ; \mathbf{s}_{t-1}; \mathbf{c}_{t'}]),
    \end{split}
\end{equation}
where $\mathbf{h}_t$ is the hidden state obtained from the Bi-LSTM module in Eq. (\ref{LSTM}), and $f^l(\cdot)$, $g(\cdot)$ are non-linear functions implemented by MLP. 
Note that $\mathbf{s}_{t-1}$ may be a state either from the high-level RL process or the low-level one.
\\
\textbf{Policy}: The stochastic policy for entity extraction $\pi : \mathcal{S} \to \mathcal{A}$ outputs an action distribution given intra-option state $\mathbf{s}_t^l$ and the high-level option $o_{t'}$ that launches the current subtask. 
\begin{equation}\label{action}
    a_t \sim \pi(a_t|\mathbf{s}_t^l; o_{t'}) = softmax(\mathbf{W}_\pi[o_{t'}] \mathbf{s}_t^l),
\end{equation}
where $\mathbf{W}_\pi$ is an array of $|\mathcal{R}|$ matrices.
\\
\textbf{Reward}\footnote{We only discuss the situation where $o_{t'}$ is included in $S$, i.e. the subtask option $o_{t'}$ is correctly predicted. Otherwise, all the low-level rewards are set to 0, which can be seen that the agent has done nothing with the low-level policy.}: Given the relation type $o_{t'}$, the entity tag for each word can be easily obtained by sampling actions from the policy. Therefore, an immediate reward $r_t^l$ is provided when the action $a_t$ is sampled by simply measuring the prediction error over gold-standard annotation:
\begin{equation}\label{reward_low}
    r_t^l = \lambda(y_t) \cdot sgn( a_t^l = y_t(o_{t'}) ), 
\end{equation}
where $sgn(\cdot)$ is the sign function, and $y(o_{t'})$ is the gold-standard entity tag conditioned on the predicted relation type $o_{t'}$. Here $\lambda(y)$ is a bias weight for down-weighing non-entity tag, defined as follows:
\begin{equation}\label{bias}
    \lambda(y) = \left\{
    \begin{array}{ll}
        1, & if ~ y \not= \texttt{N}\\
        \alpha, & if ~ y = \texttt{N}.
    \end{array}
    \right.
\end{equation}
The smaller $\alpha$ leads to less reward on words that are not entities. In this manner, the model avoids to learn a trivial policy that predicts all words as \texttt{N} (non-entity words).
When all the actions are sampled, an additional final reward $r_{fin}^l$ is computed. If all the entity tags are predicted correctly, then the agent receives +1 reward, otherwise -1. 

\subsection{Hierarchical Policy Learning}

To optimize the high-level policy, we aim to maximize the expected cumulative rewards from the main task at each time step $t$ as the agent samples trajectories following the high-level policy $\mu$, which can be computed as follows: 
\begin{equation}\label{return_high}
    J(\theta_{\mu,t}) = \mathbb{E}_{\mathbf{s}^h,o,r^h \sim \mu(o|\mathbf{s}^h)} [\sum_{k=t}^T \gamma^{k-t} r_k^h], 
\end{equation}
where $\mu$ is parameterized by $\theta_\mu$, $\gamma$ is a discount factor in RL, and the whole sampling process $\mu$ takes $T$ time steps before it terminates. 

Similarly, we learn the low-level policy by maximizing the expected cumulative intra-option rewards from the subtask over option $o_{t'}$ when the agent samples along low-level policy $\pi(\cdot|o_{t'})$ at time step $t$:
\begin{equation}\label{return_low}
    \begin{split}
        J(\theta_{\pi,t};o_{t'}) =& \mathbb{E}_{\mathbf{s}^l,a,r^l \sim \pi(a|\mathbf{s}^l;o_{t'})} [\sum_{k=t}^{T'} \gamma^{k-t} r_k^l ], 
    \end{split}  
\end{equation}
if the subtask ends at time step $T'$.

By decomposing the cumulative rewards into a Bellman equation, we acquire:
\begin{equation}\label{return}
    \begin{split}
        R^\mu(\mathbf{s}_t^h, o_t) =& \mathbb{E} [\sum_{j=0}^{N-1} \gamma^j r_{t+j}^h + \\ &\gamma^{N} R^\mu(\mathbf{s}_{t+N}^h, o_{t+N}) |\mathbf{s}_t^h, o_t], \\
        R^\pi(\mathbf{s}_t^l, a_t; o_{t'}) =& \mathbb{E} [r_t^l + \gamma R^\pi(\mathbf{s}_{t+1}^l, a_{t+1}; o_{t'}) | \mathbf{s}_t^l, a_t],
    \end{split}
\end{equation}
where $N$ is the number of time steps that a subtask continues when the entity extraction policy runs upon option $o_t$
, so the agent's next option is $o_{t+N}$. In particular, if $o_t = \texttt{NR}$, then $N = 1$.

Then, we use policy gradient methods \cite{sutton2000policy} with the REINFORCE algorithm \cite{williams1992simple} to optimize both high-level and low-level policies. 
With the likelihood ratio trick, the gradient for the high-level policy yields:
\begin{equation}\label{gradient_high}
    \begin{split}
        \nabla_{\theta_{\mu}} J(\theta_{{\mu},t}) =& \mathbb{E}_{\mathbf{s}^h,o,r^h \sim \mu(o|\mathbf{s}^h)} [R^\mu(\mathbf{s}_t^h, o_t) \\ & \nabla_{\theta_{\mu}} \log \mu_{}(o|\mathbf{s}_t^h)],
    \end{split} 
\end{equation}
and the gradient for the low-level policy yields: 
\begin{equation}\label{gradient_low}
    \begin{split}
         \nabla_{\theta_{\pi}} J(\theta_{\pi,t};o_{t'}) =& \mathbb{E}_{\mathbf{s}^l,a,r^l \sim \pi(a|\mathbf{s}^l;o_{t'})} [R^\pi(\mathbf{s}_t^l, a_t; o_{t'}) \\ & \nabla_{\theta_{\pi}} \log \pi_{}(a|\mathbf{s}_t^l; o_{t'})].
    \end{split}
\end{equation}


The entire training process is described at Algorithm \ref{training}.

\begin{algorithm}[!htb]
\small
\caption{Training Procedure of HRL}
\label{training}
Calculate $\mathbf{h}_t$ for each word in the sentence with Bi-LSTM \;
Initiate state $\mathbf{s}_0^h \leftarrow \mathbf{0}$ and time step $t \leftarrow 0$\;
\For {$i\leftarrow 1$ \KwTo Text Length}{
    $t \leftarrow t + 1$ \;
    Calculate $\mathbf{s}_t^h$ by Eq. (\ref{state_high})\;
    Sample $o_t$ from $\mathbf{s}_t^h$ by Eq. (\ref{option})\;
    Obtain reward $r_t^h$ by Eq. (\ref{reward_high})\;
    \If{$o_t \neq \texttt{NR}$}{
        \For {$j\leftarrow 1$ \KwTo Text Length}{
            $t \leftarrow t + 1$ \;
            Calculate $\mathbf{s}_t^l$ by Eq. (\ref{state_low})\;
            Sample $a_t^l$ from $\mathbf{s}_t^l$ by Eq. (\ref{action})\;
            Obtain reward $r_t^l$ by Eq. (\ref{reward_low})\;
        }
        Obtain low-level final reward $r_{fin}^l$\;
    }
}
Obtain high-level final reward $r_{fin}^h$ by Eq. (\ref{reward_h_final})\;
Optimize the model with Eq. \ref{gradient_high} and Eq. (\ref{gradient_low})\;
\end{algorithm}

\section{Experiments}

\subsection{Experimental Setting}
\subsubsection{Datasets}
We evaluated our model on the New York Times corpus which is developed by distant supervision and contains \textit{noisy} relations. The corpus has two versions: 1) The original version generated by aligning the raw data with Freebase relations \cite{riedel2010modeling}; 2) A smaller version of which the test set was manually annotated \cite{hoffmann2011knowledge}. We name the original version as {\it NYT10}, and the smaller version as {\it NYT11}. We split some of the training data from \textit{NYT11} to construct \textit{NYT11-plus}, which will be described later.



We filtered the datasets by removing 1) the relations in the training set whose relation type does not exist in the test set; 2) the sentences that contain no relations at all. Such a preprocess is also in line with the settings in the literature (for instance, \textit{Tagging}). All the baselines are evaluated in this setting for fair comparison. The statistics of the two filtered datasets are presented in Table \ref{datasets}.

\begin{table}[!htb]
    \centering
    \begin{tabular}{lrr}
    \toprule
        Dataset & NYT10 & NYT11 \\
    \midrule
        \# Relation types & 29 & 12 \\
        \# Training sentences & 70,339 & 62,648 \\
        \# Training relations & 87,739 & 74,312 \\
        \# Test sentences & 4,006 & 369 \\
        \# Test relations & 5,859 & 370 \\
    \bottomrule
    \end{tabular}
    \caption{Statistics of the datasets.}
    \label{datasets}
\end{table}

For each dataset, we randomly chose 0.5\% data from the training set for validation.

\subsubsection{Parameter Settings}
All hyper-parameters are tuned on the validation set. The dimension of all vectors in Eq. (\ref{state_high}), (\ref{LSTM}) and (\ref{state_low}) is $300$. The word vectors are initialized using Glove vectors \cite{pennington2014glove} and are updated during training. Both \textit{relation type vectors} and \textit{entity tag vectors} are initialized randomly. 
The learning rate is $4e-5$, the mini-batch size is $16$, $\alpha = 0.1$ in Eq. (\ref{bias}), $\beta = 0.9$ in Eq. (\ref{reward_h_final}), and the discount factor $\gamma = 0.95$.

\subsubsection{Evaluation Metrics}
We adopted standard micro-$F_1$ to evaluate the performance. We compared whether the extracted entity mentions can be exactly matched with those in a relation.
A triplet is regarded as correct if the relation type and the two corresponding entities are all correct. 

\subsubsection{Baselines}
We chose two types of baselines: one is pipelined methods (FCM), and the other is joint learning methods which include feature-based methods (MultiR and CoType) and neural methods (SPTree, Tagging and CopyR). We used open source codes and conducted the experiments by ourselves.
\\
\textbf{FCM} \cite{gormley2015improved}: a compositional model that combines lexicalized linguistic contexts and word embeddings to learn representations for the substructures of a sentence in relation extraction\footnote{As \textit{FCM} cannot detect entity mentions alone, we used the NER results and related features obtained from another baseline \textit{CoType}.}.
\\
\textbf{MultiR} \cite{hoffmann2011knowledge}: a typical distant supervision method performing sentence-level and corpus-level extraction, which uses multi-instance weighting to deal with noisy labels in training data.
\\
\textbf{CoType} \cite{ren2017cotype}: a domain-independent framework by jointly embedding entity mentions, relation mentions, text features, and type labels into representations, which formulates extraction as a global embedding problem. 
\\
\textbf{SPTree} \cite{miwa2016end}: an end-to-end relation extraction model that represents both word sequence and dependency tree structures using bidirectional sequential and tree-structured LSTM-RNNs.
\\
\textbf{Tagging} \cite{zheng2017joint}: an approach that treats joint extraction as a sequential labeling problem using a tagging schema where each tag encodes entity mentions and relation types at the same time.
\\
\textbf{CopyR} \cite{zeng2018extracting}: a Seq2Seq learning framework with a copy mechanism for joint extraction, where multiple decoders are applied to generate triples to handle overlapping relations.

\subsection{Main Results}

\begin{table}[!htb]
    \centering
    \begin{tabular}{lcccccc}
    \toprule
        \multirow{2}{*}{Model} & \multicolumn{3}{c}{NYT10 
        } & \multicolumn{3}{c}{NYT11}\\ 
         & Prec & Rec & $F_1$ & Prec & Rec & $F_1$\\ 
    \midrule
        FCM & -- & -- & -- & .432 & .294 & .350 \\
        MultiR & -- & -- & -- & .328 & .306 & .317 \\
        CoType & -- & -- & -- & .486 & .386 & .430 \\
        SPTree & .492 & .557 & .522 & .522 & \textbf{.541} & .531 \\
        Tagging & .593 & .381 & .464 & .469 & .489 & .479 \\
        CopyR & .569 & .452 & .504 & .347 & .534 & .421 \\
    \midrule
        HRL & \textbf{.714} & \textbf{.586} & \textbf{.644} & \textbf{.538} & .538 & \textbf{.538} \\
    \bottomrule
    \end{tabular}
    \caption{Main results on relation extraction.}
    \label{main}
\end{table}

The results on relation extraction are presented in Table \ref{main}. Noticeably, there is a significant gap between the performance on noisy data (\textit{NYT10}) and that on clean data (\textit{NYT11}) as all the models are trained on noisy data. It can be seen that our method (HRL) outperforms the baselines on the two datasets. Significant improvements can be observed on \textit{NYT10}, which indicates that our method is more robust to noisy data. 
Results on \textit{NYT11} show that neural models (\textit{SPTree}, \textit{Tagging} and \textit{CopyR}) are more effective than pipelined (\textit{FCM}) or feature-based (\textit{MultiR} and \textit{CoType}) methods. 
\textit{CopyR} is introduced to extract overlapping relations, but it yields poor performance on the \textit{NYT11} test set where there is almost no overlapping relation in a sentence (370 relations among 369 sentences). Whereas our model is still comparable to \textit{SPTree} and performs remarkably better than other baselines. Note that \textit{SPTree} utilizes more linguistic resources (e.g., POS tags, chunks, syntactic parsing trees). This implies that our model is also robust to the data distribution of relations.

\subsection{Overlapping Relation Extraction}


We prepared another two test sets to verify the effectiveness of our model on extracting overlapping relations. Note that overlapping relations can be classified into two types. 
\begin{itemize}
    \item Type I: {\it two triples share only one entity within a sentence}
    \item Type II: {\it two triples share two entities (both head and tail entities) within a sentence}
\end{itemize}
The first set,
\textit{NYT11-plus}, is annotated manually and consists of 149 sentences split from the original \textit{NYT11} training data. The set contains 210/97 overlapping relations for type I/II respectively.
The second set,
\textit{NYT10-sub}, is a subset of the test set of \textit{NYT10}, and has 715 sentences, but without manual annotation. This set contains 90/2,082 overlapping relations for type I/II respectively. To summarize, most of the overlapping relations in \textit{NYT11-plus} is of type I; while most in \textit{NYT10-sub} is of type II.
Table \ref{overlapping} shows the performance of extracting overlapping relations by different approaches. 

\begin{table}[!htb]
    \centering
    \begin{tabular}{lcccccc}
    \toprule
        \multirow{2}{*}{Model} & \multicolumn{3}{c}{NYT10-sub} & \multicolumn{3}{c}{NYT11-plus}  \\
        & Prec & Rec & $F_1$ & Prec & Rec & $F_1$ \\
    \midrule
        FCM    & -- & -- & -- & .234 & .199 & .219 \\
        MultiR & -- & -- & -- & .241 & .214 & .227 \\
        CoType & -- & -- & -- & .291 & .254 & .271 \\
        SPTree & .272 & .315 & .292 & \textbf{.466} & .229 & .307 \\
        Tagging & .256 & .237 & .246 & .292 & .220 & .250 \\
        CopyR & .392 & .263 & .315 & .329 & .224 & .264 \\
    \midrule
        HRL    & \textbf{.815} & \textbf{.475} & \textbf{.600} & .441 & \textbf{.321} & \textbf{.372} \\
    \bottomrule
    \end{tabular}
    \caption{Performance comparison on extracting overlapping relations.}
    \label{overlapping}
\end{table} 

Results on \textit{NYT10-sub} show that the baselines are very weak to extract overlapping relations of type II on the noisy data, which is consistent with our statement that existing joint extraction approaches cannot deal with overlapping relations effectively in nature. By contrast, our method did not deteriorate too much in performance comparing to that in Table \ref{main}, and even obtained a larger gain on \textit{precision}. 


Results on \textit{NYT11-plus} demonstrate that our method had a substantial $F_1$ improvement over all the baselines in extracting overlapping relations of type I on the clean data, indicating that our method can extract overlapping relations more accurately. \textit{SPTree} had a high \textit{precision} but low \textit{recall} since it simply matches one relation type to an entity pair, suffering from ignoring the case of overlapping relations. \textit{Tagging} had low performance in extracting overlapping relations because it assigns a unique tag to an entity even if that entity participates in overlapping relations. Though \textit{CopyR} claimed that it can extract overlapping relations of both types, it fails to extract the relations from clean data effectively as it strongly relies on the annotation of the noisy training data.

To conclude, we can see that extracting overlapping relations is more challenging by comparing results in Table \ref{main} and those in Table \ref{overlapping}, and our model is better in extracting two types of overlapping relations no matter the data is noisy or clean. 


\begin{table*}[!htb]
    \centering
    \begin{tabularx}{\textwidth}{X}
        \toprule
        The lawsuit contended that the chairman of the {\bf\normalsize\color{brown}[} {\bf\normalsize\color{red}[} News Corporation {\bf\normalsize\color{red}]${}_\text{Et-Company}$} {\bf\normalsize\color{brown}]${}_\text{Es-Founder}$} , {\bf\normalsize\color{blue}[}  {\bf\normalsize\color{brown}[} {\bf\normalsize\color{red}[} Rupert {\xmybox[red]{Murdoch}} {\bf\normalsize\color{red}]}{\bf\normalsize\color{red}${}_\text{Es-Company}$} {\bf\normalsize\color{brown}]${}_\text{Et-Founder}$} {\bf\normalsize\color{blue}]${}_\text{Es-Nationality}$}  {\xmybox[brown]{,{\color{brown!10!white}l}}} promised certain rights to shareholders , including the vote on the poison pill , in return for their approval of the company 's plan to reincorporate in the United States from {\bf\normalsize\color{blue}[} {\xmybox[blue]{Australia}} {\bf\normalsize\color{blue}]$_\text{Et-Nationality}$} .  \\
        \midrule
        Both {\bf\normalsize\color{red}[} Steven A. Ballmer {\bf\normalsize\color{red}]${}_\text{Es-Company}$}  , {\bf\normalsize\color{blue}[} {\normalsize\color{brown}[} {\bf\normalsize\color{red}[} {\xmybox[red]{Microsoft} } {\bf\normalsize\color{red}]${}_\text{Et-Company}$} {\bf\normalsize\color{brown}]${}_\text{Et-Company}$} {\bf\normalsize\color{blue}]$_\text{Es-Founder}$} 's chief executive , and {\bf\normalsize\color{blue}[} {\bf\normalsize\color{brown}[} Bill {\xmybox[brown]{Gates}} {\bf\normalsize\color{brown}]${}_\text{Es-Company}$} {\bf\normalsize\color{blue}]$_\text{Et-Founder}$} {\xmybox[blue]{,{\color{blue!10!white}l}}} the chairman , have been involved in that debate inside the company , according to that person .\\
        \bottomrule
    \end{tabularx}
    \caption{Extraction examples by our model. The words in a bracket represents an entity extracted by the model. \textit{Es} stands for source entity and \textit{Et} for target entity. A predicted relation indicator is marked in background color (e.g. ``Murdoch" in the first instance). The entities which form a triple are bracketed in the same color.}
    \label{case}
\end{table*}

\subsection{Interaction between the Two Policies}

To justify the effectiveness of integrating entities into a relation and how the interactions are built between the two policies, we investigated the performance on relation detection (classification)
. In this setting, a prediction is treated as correct as long as the relation type is correctly predicted. The prediction is derived from the high-level policy. 

\begin{table}[!htb]
    \centering
    \begin{tabular}{lcccccc}
    \toprule
        \multirow{2}{*}{Model} & \multicolumn{3}{c}{NYT11} & \multicolumn{3}{c}{NYT11-plus}\\
         & Prec & Rec & $F_1$ & Prec & Rec & $F_1$ \\
    \midrule
        FCM    & .502 & .479 & .490 & .447 & .327 & .378 \\
        MultiR & .465 & .439 & .451 & .423 & .336 & .375 \\
        CoType & .558 & .558 & .558 & .491 & .413 & .449 \\
        SPTree & .650 & .614 & .631 & \textbf{.700} & .343 & .460 \\
        CopyR & .480 & \textbf{.714} & .574 & .626 & .426 & .507\\
    \midrule
        HRL-Ent  & \textbf{.676} & .676 & \textbf{.676} & .577 & .321 & .413 \\
        HRL    & .654 & .654 & .654 & .626 & \textbf{.456} & \textbf{.527} \\
    \bottomrule
    \end{tabular}
    \caption{Performance comparison on relation detection.} 
    \label{relation}
\end{table} 

The results in Table \ref{relation} demonstrate that our method performs better in relation detection on both datasets. The improvements on {\it NYT11-plus} are more remarkable as our paradigm is more powerful to extract multiple relations from a sentence. 
The results indicate that our extraction paradigm which regards entities as arguments of a relation can better capture the relational information in the text.

When removing the low-level entity extraction policy from our model (HRL-Ent), the performance has changed slightly on \textit{NYT11} because each sentence almost contains only one relation in this test set (370 relations among 369 sentences). 
In this case, the interaction between the two policies has almost no influence on relation detection. However, dramatic drops are observed on \textit{NYT11-plus} where we have 327 relations from 149 sentences, implying that our method (HRL) captures the dependency across multiple extraction tasks and the high-level policy benefits from such interactions. Therefore, our hierarchical extraction framework indeed enhances the interaction between relation detection and entity extraction.

\subsection{Case Study}


Table \ref{case} presents some extraction examples by our model to demonstrate the ability to extract overlapping relations. The first sentence shows the case that an entity pair has multiple relations (type II). Two relations (\textit{Rupert Murdoch}, \texttt{person-company}, \textit{News Corporation}) and (\textit{News Corporation}, \texttt{company-founder}, \textit{Rupert Murdoch}) share the same entity pair but have different relation types. The model first detects the relation type \texttt{person-company} at ``Murdoch", and then detects the other relation type \texttt{company-founder} at the \textit{comma} position, just next to the word ``Murdoch". This shows that relation detection is triggered when sufficient evidence has been gathered at a particular position. 
And the model can classify the same entities into either source or target entities (for instance, \textit{Rupert Murdoch} is a source entity for \texttt{person-company} whereas a target entity for \texttt{company-founder}), demonstrating the advantage of our hierarchical framework which can assign dynamic tags to words conditioned on different relation types.
In addition, \textit{Rupert Murdoch} has a relation with \textit{Australia}, where the two entities locate far from each other. Though this is more difficult to detect, our model can still extract the relation correctly.

The second sentence gives another example where an entity is involved in multiple relations (type I). In this sentence, (\textit{Steven A. Ballmer}, \texttt{person-company}, \textit{Microsoft}) and (\textit{Bill Gates}, \texttt{person-company}, \textit{Microsoft}) share the same relation type and target entity, but have different source entities. When the agent scans to the word ``Microsoft", the model detects the first relation. The agent then detects the second relation when it scans to the word ``Gates". This further demonstrates the benefit of our hierarchical framework which has strengths in extracting overlapping relations by firstly detecting relation and then finding the entity arguments. 
In addition, our model predicts another relation (\textit{Bill Gates}, \texttt{founder-of}, \textit{Microsoft}), which is wrong for this sentence because there is no explicit mention of the relation. This may result from the noise produced by distant supervision, where there are many noisy sentences that are aligned to that relation.

\section{Conclusion and Future Work}
In this paper, we present a hierarchical extraction paradigm which approaches relation extraction via hierarchical reinforcement learning.
The paradigm treats entities as the arguments of a relation, and decomposes the relation extraction task into a hierarchy of two subtasks: high-level relation indicator detection and low-level entity mention extraction. The high-level policy for relation detection identifies multiple relations in a sentence, and the low-level policy for entity extraction launches a subtask to further extract the related entities for each relation. 
Thanks to the nature of this hierarchical approach, it is good at modeling the interactions between the two subtasks, and particularly excels at extracting overlapping relations. 
Experiments demonstrate that our approach outperforms state-of-the-art baselines.

As future work, this hierarchical extraction framework can be generalized to many other pairwise or triple-wise extraction tasks such as aspect-opinion mining or ontology induction.

\section*{Acknowledgements}
This work was jointly supported by the National Science Foundation of China  (Grant No.61876096/61332007), and the National Key R\&D Program of China (Grant No. 2018YFC0830200). We would like to thank Prof. Xiaoyan Zhu for her generous support.

\bibliographystyle{aaai}
\bibliography{aaai19}

\end{document}